\pdfoutput=1

\documentclass[11pt]{article}
\usepackage{booktabs}
\usepackage{marvosym}
\usepackage{multirow}
\usepackage{subcaption}
\usepackage{array}
\usepackage{calc}
\usepackage{algorithm} 
\usepackage{algpseudocode} 
\usepackage{xcolor}

\usepackage[preprint]{acl}
\usepackage{amsmath}
\usepackage{times}
\usepackage{latexsym}
\usepackage{amssymb}
\usepackage[T1]{fontenc}

\usepackage[utf8]{inputenc}

\usepackage{microtype}

\usepackage{inconsolata}

\usepackage{graphicx}

\usepackage{enumitem} 
\setlist[itemize]{noitemsep, topsep=0pt} 

\title{EdgeInfinite: A Memory-Efficient Infinite-Context Transformer for \\ Edge Devices}

\author{
 \textbf{Jiyu Chen\textsuperscript{*}\textsuperscript{1,2}},
 \textbf{Shuang Peng\textsuperscript{*}\textsuperscript{1}},
 \textbf{Daxiong Luo\textsuperscript{*}\textsuperscript{1}},
 \textbf{Fan Yang\textsuperscript{1}},
 \textbf{Renshou Wu\textsuperscript{1}},
 \\
 \textbf{Fangyuan Li\textsuperscript{1\Letter}},
 \textbf{Xiaoxin Chen\textsuperscript{1}}
\\
 \textsuperscript{1}vivo AI Lab,
 \textsuperscript{2}Zhejiang University
\\
\textsuperscript{*}Equal contribution
\textsuperscript{\Letter}Corresponding author
\\
jiyuchen@zju.edu.cn, \{pengshuang,luodaxiong,lifangyuan\}@vivo.com
}

\begin{document}
\maketitle
\begin{abstract}
Transformer-based large language models (LLMs) encounter challenges in processing long sequences on edge devices due to the quadratic complexity of attention mechanisms and growing memory demands from Key-Value (KV) cache. Existing KV cache optimizations struggle with irreversible token eviction in long-output tasks, while alternative sequence modeling architectures prove costly to adopt within established Transformer infrastructure. We present EdgeInfinite\footnote {The code will be released after the official audit.}, a memory-efficient solution for infinite contexts that integrates compressed memory into Transformer-based LLMs through a trainable memory-gating module. This approach maintains full compatibility with standard Transformer architectures, requiring fine-tuning only a small part of parameters, and enables selective activation of the memory-gating module for long and short context task routing. The experimental result shows that EdgeInfinite achieves comparable performance to baseline Transformer-based LLM on long context benchmarks while optimizing memory consumption and time to first token. 
\end{abstract}

\section{Introduction}

The Transformer \cite{vaswani2017attention} has become the foundational framework for Large Language Models (LLMs). However, the quadratic time complexity of the classic attention mechanism in Transformer-based model presents significant challenges in processing long sequences. Moreover, the continuous growth of the Key-Value (KV) cache, driven by increasing context lengths, leads to increased memory usage. Whether in terms of time complexity or limited memory, these challenges are particularly pronounced on resource-constrained edge devices such as smartphones.

To address these challenges, two main solutions have been proposed. One approach focuses on the KV cache optimizations \cite{li2024snapkv,xiao2023efficient,zhang2023h2o}, primarily by evicting tokens deemed unimportant to reduce attention computation complexity. Though these methods can improve efficiency, they may encounter a potential issue that the evicted tokens will not be used in the future \cite{tang2024razorattention}, especially in real-world scenarios, such as multi-round interactions \cite{li2024scbench, qin2024kimi} and long-generation Chain-of-Thought (CoT) reasoning \cite{wei2022chain, guo2025deepseek}.

The second solution explores more efficient sequence modeling methods, such as linear recurrent models \cite{katharopoulos2020transformers, li2025minimax} and state space models \cite{gu2021efficiently, gu2023mamba}, to address computational complexity issues. However, most current work remains centered around Transformer-based models. Adopting new structural models would incur substantial costs, hindering their deployment on edge devices.

In this work, we propose \textbf{EdgeInfinite}, a novel approach that efficiently handles long sequences on edge devices. By continuing pre-training with existing Transformer-based LLMs, EdgeInfinite maintains compatibility with current Transformer architecture, enabling a more streamlined and resource-efficient approach to model development. We design a trainable memory-gating module that requires fine-tuning only a small subset of parameters. This module can be selectively loaded for long text tasks, while retaining the original parameters of the Transformer model for short text tasks. This flexibility ensures that the base model's parameters do not require additional fine-tuning, allowing for rapid and efficient inference on long text tasks. As a result, our approach is well-suited for deployment on edge devices. During inference, we retain sink tokens and window tokens in KV cache, while the other KV pairs are compressed into the memory block. This approach allows the model to preserve more semantic and positional information during inference. Moreover, EdgeInfinite demonstrates the improvement in time to first token (TTFT), a notable advancement among existing methods.

Our contributions can be summarized as follows:

\begin{itemize}[itemsep=1pt,topsep=1pt,parsep=1pt,leftmargin=0.5cm]
    \item We propose EdgeInfinite, an edge-side infinite context method that integrates compressed memory with a trainable memory-gating module, while maintaining compatibility with the vanilla Transformer architecture.
    \item EdgeInfinite maintains the original Transformer-based LLM's performance on short text tasks while supporting high-efficiency inference for long text tasks. This mechanism is highly suitable for model deployment on resource-constrained edge devices.
    \item We evaluate the performance of EdgeInfinite on long context benchmark. It achieves performance comparable to the baseline Transformer-based models while optimizing memory consumption and TTFT.
\end{itemize}

\section{Related work}

The quadratic time complexity of the attention mechanism and the growing memory use of the KV cache in classic Transformer-based LLMs pose challenges for processing long sequences on resource-constrained edge devices. This section highlights recent work to address these issues.

\noindent\textbf{Innovative Sequence Models} Mamba \cite{gu2023mamba} and Mamba-2 \cite{dao2024transformers} represent the significant milestone in the development of State Space Model (SSM) \cite{gu2021efficiently}, demonstrating outstanding performance in natural language processing and other tasks. The RWKV \cite{peng2023rwkv,peng2024eagle} combines the advantages of RNN and Transformer, introducing innovations such as token shift and optimized time-mixing to achieve linear complexity in inference. Titans \cite{behrouz2024titans} combine attention as short-term memory with a neural long-term memory module. Infini-Transformer \cite{munkhdalai2024leave} segments long sequences into multiple blocks, incorporates a compressive memory into the vanilla attention mechanism and builds in both masked local attention and long-term linear attention mechanisms in a single Transformer block.

\noindent\textbf{KV cache Optimizations} KV Cache Optimizations primarily aim to reduce overall computational requirements by identifying and discarding unimportant tokens. StreamingLLM \cite{xiao2023efficient} is a method based on sliding window attention. By retaining both the most recent and sink tokens, it helps maintain the model's performance while efficiently managing memory usage. H2O \cite{zhang2023h2o} employs attention scores to identify and retain significant tokens while simultaneously preserving the most recent tokens. SnapKV \cite{li2024snapkv} identifies critical attention features based on observation windows and correspondingly compresses the KV cache. PyramidKV \cite{cai2024pyramidkv} reduces the KV cache budget for later layers by analyzing the attention features across different layers. SCOPE \cite{wu2024scope} innovatively refines the KV cache budget problem by considering it separately in the prefill and decode stages.

\begin{figure*}[t]
    \centering
  \includegraphics[width=0.8\textwidth]{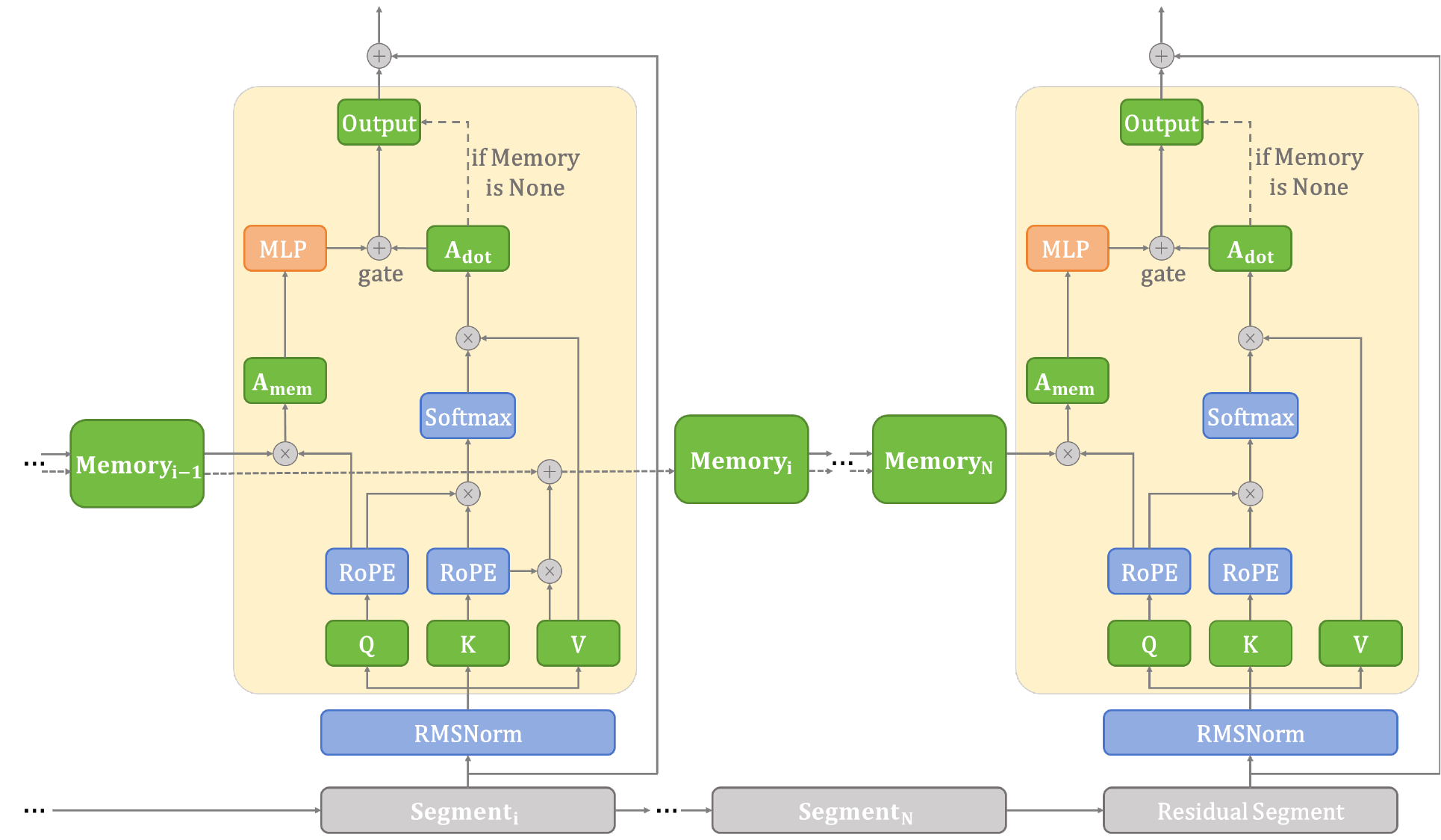}
  \caption{The overall framework of EdgeInfinite: illustrating the computation process of the attention layer in Transformer-based LLMs, with LLaMA Attention \cite{touvron2023llama,grattafiori2024llama} as an example.}
  \label{fig:framework}
\end{figure*}

\section{EdgeInfinite}
\subsection{Architecture}
As shown in Figure~\ref{fig:framework}, the architecture of EdgeInfinite includes three core components: (1) Segmented attention with Rotary Position Embedding (ROPE) for local context modeling, (2) The memory mechanism for compressing and decompressing historical context, and (3) The adaptive memory-gating module that balances local and memory-based attention.

\subsubsection{Segmented Attention with ROPE}
Given an input sequence $ X = [x_1, \ldots, x_L]^T \in \mathbb{R}^{L \times d} $, it is divided into segments of size $ L_{seg} $, resulting in $ N $ segments of length $ L_{seg} $ and a residual segment of length $ L_{res} $. Their relationship can be expressed as:
\begin{equation}
    \label{eq:formula1}
    L = N \cdot L_{seg} + L_{res}
\end{equation}

The full segment $ X_{seg} \in \mathbb{R}^{L_{seg} \times d} $ or the residual segment $ X_{res} \in \mathbb{R}^{L_{res} \times d} $ can be collectively represented as $ X_{s/r} \in \mathbb{R}^{L_{s/r} \times d} $, where $s/r$ indicates either a full or residual segment. We compute the attention query $ Q $, key $ K $, and value $ V $ states:
\begin{equation}
    \label{eq:formula2}
    Q = X_{s/r} W^Q,K = X_{s/r} W^K,V = X_{s/r} W^V
\end{equation}
where $ W^K $, $ W^V $, and $ W^Q $ are the trainable projection matrices. $Q=[q_1, q_2, \ldots, q_{L_{s/r}}]$ and $K = [k_1, k_2, \ldots, k_{L_{s/r}}]$ denote the query and key states in the segment $X_{s/r}$, where $q_i$ and $k_i$ represent the query and key states corresponding to the $i$-th token. 

Next, the ROPE model \cite{su2024roformer} is integrated to incorporate positional information into the attention computation:
\begin{equation}
    \label{eq:formula3}
    q_m^{r}=\mathcal{R}_mq_m, k_n^{r}=\mathcal{R}_nk_n
\end{equation}
where $R_m$ and $R_n$ are the rotary matrices situated at positions $m$ and $n$. $q_m^{r}$ and $k_n^{r}$ represent the query and key states after the rotary transformation. After applying the rotary transformation, the modified query and key states are denoted as $Q^{r}$ and $K^{r}$. 

Subsequently, the attention computation for each segment is performed in a manner similar to the vanilla Transformer architecture \cite{vaswani2017attention}:
\begin{equation}
    \label{eq:formula4}
    A_{\mathrm{dot}} = \mathrm{softmax}(\frac{Q^{r}(K^{r})^T}{\sqrt{d}})V
\end{equation}
This computation enables the model to capture dependencies between tokens within each segment while incorporating positional information through the ROPE model.

\subsubsection{Memory Compression-Decompression}
Inspired by the Infini-Transformer \cite{munkhdalai2024leave} and linearized attention \cite{katharopoulos2020transformers}, we introduce memory compression and memory decompression. For all segments except the residual segment, memory compression is performed. For the $i$-th segment, the memory $M_{i}$ and the normalization term $z_i$ are calculated as follows:
\begin{equation}
    \label{eq:formula5}
    M_{i} = M_{i-1}+\sigma (K^{r})^TV
\end{equation}
\begin{equation}
    \label{eq:formula6}
    z_{i} = z_{i-1}+\sum_{j=1}^{L_{s/r}}\sigma (k_j^{r})
\end{equation}
where $\sigma$ denotes a nonlinear activation function. $M_{i} \in \mathbb{R}^{d \times d}$ and $z_i \in \mathbb{R}^{d \times 1}$ are both initialized as zero matrices for the first segment ($i=1$). Here, the memory $M_i$ stores the associations between the keys and values of previous segments. The nonlinear activation function and normalization are primarily used to ensure the stability of model training.

For all segments, the memory decompression is executed as follows:
\begin{equation}
    \label{eq:formula7}
    A_{\mathrm{mem}} = \frac{\sigma(Q^{r})M_{i-1}}{\sigma(Q^{r})z_{i-1}}
\end{equation}
where $A_{\mathrm{mem}}\in \mathbb{R}^{L_{r/s} \times d}$ represents the attention calculated by the memory and query state of the current segment. Since the memory encodes the associations of key-value pairs from previous segments, decompression allows us to compute the attention between the current query state and the past key-value states. This process enables blockwise computation to approximate the attention calculation of the original long sequence.

\subsubsection{Memory-Gating Module}
In contrast to the Infini-Transformer, which requires training the entire model, our approach requires fine-tuning only the memory-gating module. This module can integrate memory-based attention with local segment-based attention, enhancing the model's ability to handle long-range dependencies. Additionally, our method supports switching to the original model for inference on short context tasks. 

The memory-gating module is a trainable component that consists of a Multi-Layer Perceptron (MLP) and a gating vector. Specifically, the memory attention $A_{\mathrm{mem}}$ is first transformed through the MLP as follows:
\begin{equation}
    \label{eq:formula8}
    \tilde{A}_{\mathrm{mem}} = W_2 \cdot \mathrm{ReLU}(W_1 A_{\mathrm{mem}} + b_1) + b_2
\end{equation}
Here, $W_1$ and $W_2$ are trainable weight matrices, while $b_1$ and $b_2$ are bias vectors. The ReLU activation function introduces non-linearity, enabling the MLP to refine the memory-based attention and capture complex interactions between the current segment and accumulated memory.

The transformed memory attention $\tilde{A}_{\mathrm{mem}}$ is then combined with the local segment-based attention ${A}_{\mathrm{dot}}$ through a gating mechanism. The combined attention $A_{\mathrm{com}}$ is computed as:
\begin{equation}
    \label{eq:formula9}
    \begin{aligned}
    A_{\mathrm{com}} &= \mathrm{sigmoid}(g) \odot \tilde{A}_{\mathrm{mem}} \\
      &\quad + (1 - \mathrm{sigmoid}(g)) \odot A_{\mathrm{dot}}
    \end{aligned}
\end{equation}
where $g$ is a trainable gating vector. The sigmoid function applied to $g$ produces a gating factor that adaptively controls the contribution of $\tilde{A}_{\mathrm{mem}}$ and $A_{\mathrm{dot}}$ to the combined attention. This adaptive weighting mechanism ensures that the model can dynamically balance the importance of previous context (encoded in $\tilde{A}_{\mathrm{mem}}$) and current context (encoded in $A_{\mathrm{dot}}$) based on the specific features of the long sequence. 

The memory-gating module is integrated as a bypass in the attention pipeline. If the sequence length is insufficient to be divided into segments, the memory is None and the memory mechanism is bypassed, reverting to standard Multi-Head Attention. The final attention output $O$ is given by:
\begin{equation}
    \label{eq:formula10}
    \begin{cases} 
    O = [A_{\mathrm{com}}^1;\dots A_{\mathrm{com}}^H] W_o & \text{if Memory} \neq \text{None} \\
    O = [A_{\mathrm{dot}}^1;\dots A_{\mathrm{dot}}^H] W_o & \text{if Memory} = \text{None}
    \end{cases}
\end{equation}
where $A_{\mathrm{com}}^h$ and $A_{\mathrm{dot}}^h$ represent the combined attention and the local segment-based attention for the $h$-th head. This design ensures consistency with the base model for short context tasks, avoiding catastrophic forgetting.

\begin{figure}[t]
  \includegraphics[width=\columnwidth]{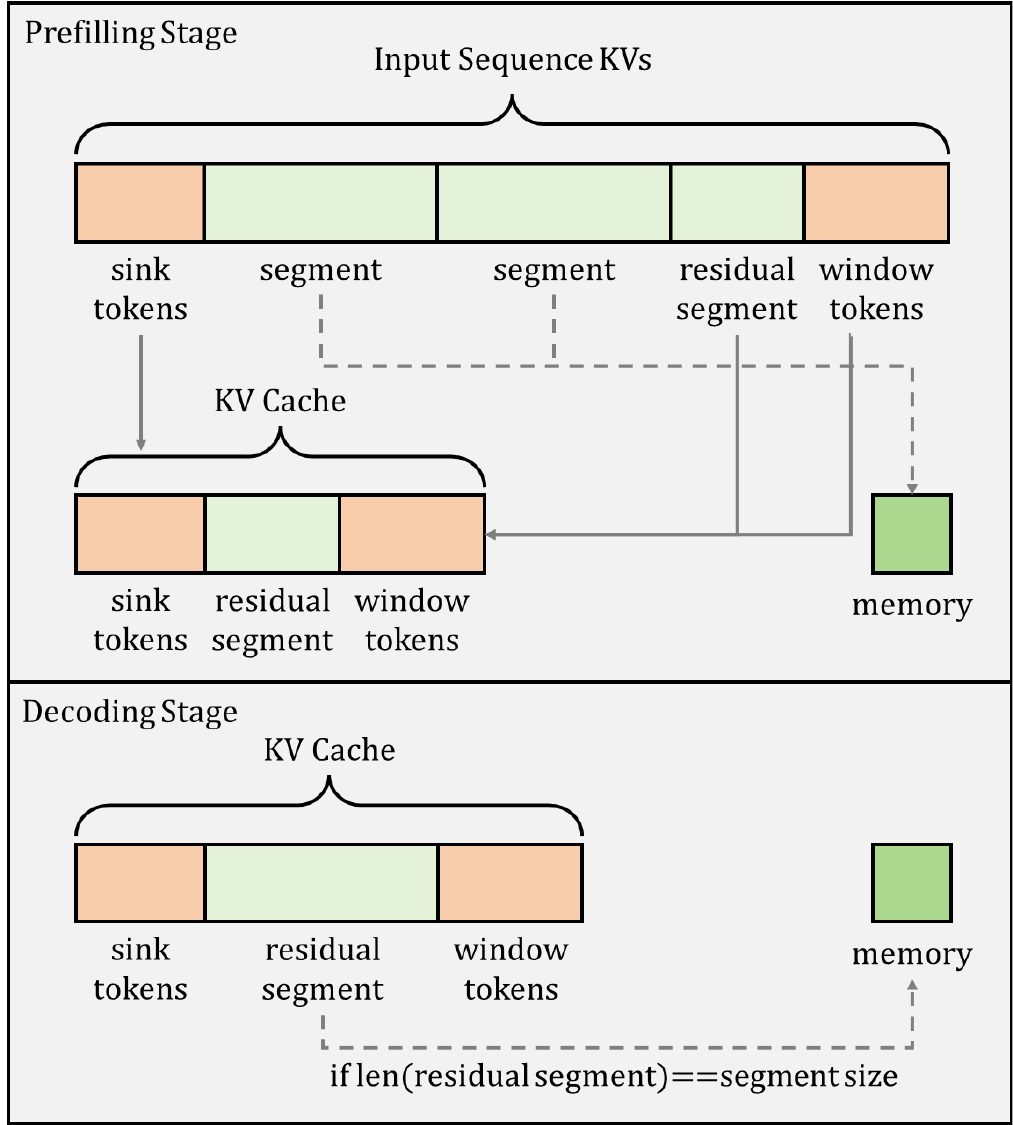}
  \caption{The inference strategy of EdgeInfinite.}
  \label{fig:inference}
\end{figure}

\begin{algorithm}[t]
\caption{EdgeInfinite Inference Strategy.}
\label{alg:inference}
\begin{algorithmic}[1]
\fontsize{7.8pt}{10pt}\selectfont
\State \textbf{Input:} Input sequence $X_{\text{in}} = [x_1, \ldots, x_L]^T$, memory $M$, normalization term $z$, KV cache $C$
\State \textbf{Output:} Output sequence $X_{\text{out}} = [x_1, \ldots, x_{L_{\text{max}}}]^T$
\State \textbf{// Prefilling stage:}
\State Initialize memory $M$, normalization term $z$, and KV cache $C$
\If{$L \geq L_{\text{sink}} + L_{\text{window}} + L_{\text{seg}}$}
    \State $C = \text{get\_kv\_cache}(X_{\text{in}}[:{L_{\text{sink}}}],C)$
    \State $N = \lfloor (L - L_{\text{sink}} - L_{\text{window}}) / L_{\text{seg}} \rfloor$
    \For{$i = 0$ \textbf{to} $N-1$}
        \State $X_{\text{segment}} = X_{\text{in}}[L_{\text{sink}} + i \cdot L_{\text{seg}} : L_{\text{sink}} + (i+1) \cdot L_{\text{seg}}]$
        \State $M, z = \text{get\_memory}(X_{\text{segment}}, C, M, z)$
    \EndFor
    \State $X_{\text{remaining}} = X_{\text{in}}[L_{\text{sink}} + N \cdot L_{\text{seg}} :]$
    \State $O, C = \text{get\_model\_output}(X_{\text{remaining}}, C, M, z)$
\Else
    \State $O, C = \text{get\_model\_output}(X_{\text{in}}, C, M, z)$
\EndIf
\State $x_{\text{new}} = \text{get\_model\_decode}(O)$
\State $X_{\text{out}} = [X_{\text{in}}; x_{\text{new}}]$
\State \textbf{// Decoding stage:}
\While{$\text{len}(X_{\text{out}}) < L_{\text{max}}$}
    \State $L_{\text{res}} = \text{len}(X_{\text{out}}) - L_{\text{sink}} - L_{\text{window}}$
    \If{$L_{\text{res}} == L_{\text{seg}}$}
        \State $X_{\text{segment}} = X_{\text{out}}[-L_{\text{seg}} - L_{\text{window}} : -L_{\text{window}}]$
        \State $C = C[:L_{\text{sink}}]$
        \State $M, z = \text{get\_memory}(X_{\text{segment}}, C, M, z)$
        \State $O, C = \text{get\_model\_output}(X_{\text{out}}[-L_{\text{window}}:], C, M, z)$
    \Else
        \State $O, C = \text{get\_model\_output}(X_{\text{out}}[-1:], C, M, z)$
    \EndIf
    \State $x_{\text{new}} = \text{get\_model\_decode}(O)$
    \State $X_{\text{out}} = [X_{\text{out}}; x_{\text{new}}]$
\EndWhile
\end{algorithmic}
\end{algorithm}

\subsection{Inference Strategy}
The inference strategy of EdgeInfinite is formalized in Algorithm~\ref{alg:inference} and visualized in Figure~\ref{fig:inference}. It is characterized by two main components: (1) Selective token preservation to ensure high-quality inference, and (2) Adaptive long-short text routing for handling of diverse input lengths.

\subsubsection{Selective Token Preservation}
EdgeInfinite significantly compresses the key states and value states associated with multiple tokens, similar to KV cache optimization methods that discard several tokens to reduce computational overhead. However, this approach may potentially degrade overall inference performance.

To address this issue, EdgeInfinite preserves two types of important tokens in the KV cache during the inference process: \textbf{sink tokens} and \textbf{window tokens}. Sink tokens represent the initial tokens of the sequence, while window tokens correspond to the most recent tokens. These tokens are crucial for preserving semantic and positional information \cite{xiao2023efficient}, and they are retained uncompressed to ensure high-quality inference outputs. 

\subsubsection{Long-Short Text Inference Routing}
EdgeInfinite's inference strategy adapts dynamically to handle both long and short text inputs efficiently. The entire inference process can be divided into prefilling stage and decoding stage:

\noindent\textbf{Prefilling Stage} For long input sequences ($ L \geq L_{\mathrm{sink}} + L_{\mathrm{window}} + L_{\mathrm{seg}} $), the sequence excluding the sink tokens and window tokens is divided into $N$ chunks, each of length $L_{seg}$. Each chunk is compressed into memory, with sink tokens concatenated in front. The remaining parts, including the residual segment, are stored as KV cache. For short input sequences ($ L < L_{\mathrm{sink}} + L_{\mathrm{window}} + L_{\mathrm{seg}} $), the model retains the full KV cache, similar to the original attention. Here, $ L_{\mathrm{sink}} $ and $ L_{\mathrm{window}} $ are the lengths of retained sink tokens and window tokens.

\noindent\textbf{Decoding Stage} The model iteratively generates new tokens until the length of the output sequence reaches $L_{max}$. If the length of the residual sequence equals $L_{\mathrm{seg}}$, the memory is updated by compressing the corresponding segment, with the sink tokens concatenated in front. The output is then generated based on the updated memory, the KV cache of sink tokens, and the KV states of window tokens. Otherwise, the model directly generates the next token using the current KV cache and memory.

\begin{table*}[t]
\centering
\begin{subtable}[t]{\textwidth}
\centering
{
\fontsize{6.5}{8.5}\selectfont
\setlength{\tabcolsep}{2.3pt} 
\setlength{\aboverulesep}{0.5pt} 
\setlength{\belowrulesep}{0.5pt} 
\begin{tabular}{cccccccccccccccc}
\hline
 & \multicolumn{5}{c}{Single-Document QA} & \multicolumn{5}{c}{Multi-Document QA} & \multicolumn{5}{c}{Summarization} \\ 
 \cmidrule(lr){2-6} \cmidrule(lr){7-11} \cmidrule(lr){12-16}
 & NrtvQA & Qasper & MF-en & MF-zh & Avg & HotpotQA & 2WikiMQA & MuSiQue & DuReader & Avg & GovReport & QMSum & MultiNews & VCSUM & Avg \\ \hline
FullKV & 5.94 & 31.50 & 34.89 & 47.88 & 30.05 & 21.93 & 26.15 & 2.58 & 24.91 & 18.89 & 12.82 & 7.04 & 10.94 & 18.34 & 12.29 \\ \hline
SnapKV & 5.53 & 29.80 & 35.04 & \textbf{48.97} & 29.84 & 22.51 & 26.04 & 2.14 & 22.77 & 18.37 & 11.09 & 6.68 & \textbf{11.08} & 17.74 & 11.65 \\
PyramidKV & 5.01 & \textbf{30.06} & \textbf{35.50} & 48.82 & \textbf{29.85} & 22.25 & 25.76 & 2.22 & 22.95 & 18.30 & 11.27 & 6.53 & 10.93 & 17.60 & 11.58 \\
StreamingLLM & 3.70 & 25.54 & 29.45 & 43.15 & 25.46 & 16.63 & 19.13 & 2.25 & 23.61 & 15.41 & 10.84 & 5.27 & 10.50 & 17.39 & 11.00 \\
EdgeInfinite & \textbf{14.16} & 18.68 & 25.58 & 35.56 & 23.50 & \textbf{31.67} & \textbf{26.08} & \textbf{12.06} & \textbf{26.87} & \textbf{24.17} & \textbf{11.28} & \textbf{8.18} & 10.76 & \textbf{18.18} & \textbf{12.10} \\
\hline
\end{tabular}
}
\caption{\label{longbench1}
Results on single-document QA, multi-document QA, and summarization tasks.}
\end{subtable}

\begin{subtable}[t]{\textwidth}
\centering
{
\fontsize{6.5}{8.5}\selectfont
\setlength{\tabcolsep}{6pt} 
\setlength{\aboverulesep}{0.5pt} 
\setlength{\belowrulesep}{0.5pt} 
\begin{tabular}{cccccccccccccccc}
\hline
  & \multicolumn{5}{c}{Few-shot Learning} & \multicolumn{4}{c}{Synthetic} & \multicolumn{3}{c}{Code} & Overall \\
    \cmidrule(lr){2-6} \cmidrule(lr){7-10} \cmidrule(lr){11-13} \cmidrule(lr){14-14}
 & TREC & TriviaQA & SAMSum & LSHT & Avg & PCount & PRe-en & PRe-zh & Avg & LCC & RB-P & Avg & Avg \\ \hline
FullKV & 63.00 & 51.98 & 24.50 & 18.00 & 39.37 & 2.50 & 4.50 & 28.00 & 11.67 & 42.96 & 27.81 & 35.39 & 24.20 \\ \hline
SnapKV & 60.00 & 51.98 & 24.32 & 17.75 & 38.51 & 1.79 & 5.50 & \textbf{30.00} & \textbf{12.43} & 43.72 & 27.07 & 35.40 & 23.88 \\
PyramidKV & \textbf{61.00} & 51.46 & 24.07 & 18.00 & 38.63 & 2.17 & 5.31 & 28.50 & 11.99 & \textbf{43.86} & 26.74 & 35.30 & 23.81 \\
StreamingLLM & \textbf{61.00} & 38.20 & 10.92 & 14.17 & 31.07 & 2.60 & 4.29 & 7.50 & 4.80 & 33.49 & 22.66 & 28.08 & 19.16 \\
EdgeInfinite & 55.00 & \textbf{79.03} & \textbf{33.27} & \textbf{24.25} & \textbf{47.89} & \textbf{3.50} & \textbf{6.00} & 24.00 & 11.17 & 42.66 & \textbf{33.09} & \textbf{37.88} & \textbf{25.71} \\ \hline
\end{tabular}
}
\caption{\label{longbench2}
Results on few-shot learning, synthetic, code tasks, and overall LongBench task average results.}
\end{subtable}

\caption{\label{longbench}
Performance comparison of EdgeInfinite (Ours) with SnapKV, PyramidKV, StreamingLLM and FullKV on LongBench.}
\end{table*}

\section{Experiments}

\subsection{Experimental Setups}

\noindent\textbf{Model and Data} In our experiments, we evaluate EdgeInfinite using BlueLM-3B \cite{lu2024bluelm} as the backbone, a Transformer-based LLM suitable for edge deployment. The training dataset includes approximately 100,000 samples, covering diverse tasks such as text summarization and generation.

\noindent\textbf{Hyperparameters} The model is trained for 2 epochs with a learning rate set to 0.005. Only the memory-gating module (0.15\% of total weights) is trained. We configure other hyperparameters as follows: $L_{\mathrm{seg}}$ is set to 2048, $ L_{\mathrm{sink}} $ to 300, and $ L_{\mathrm{window}} $ to 200. For sequences of varying lengths, the total size of the retained KV cache averages approximately 1524 tokens, which includes 300 sink tokens, 200 window tokens, and an average residual segment length of 1024 tokens.

\noindent\textbf{Benchmark} We evaluate EdgeInfinite using LongBench \cite{bai2023longbench}, a multi-task long-context benchmark for assessing long-context comprehension abilities across diverse datasets. 

\noindent\textbf{Baseline} We compare EdgeInfinite with three baseline KV cache optimization methods, including SnapKV \cite{li2024snapkv}, PyramidKV \cite{cai2024pyramidkv}, and StreamingLLM \cite{xiao2023efficient}, as well as the original model with full KV cache. The cache sizes for these three baselines are set to 2048, slightly larger than the setting of EdgeInfinite.

\subsection{Results}

The performance comparison between baseline and our method is shown as Table~\ref{longbench}. We report the average performance for each category , as well as the overall average performance across all 21 tasks. 

Overall, EdgeInfinite demonstrates competitive performance advantages compared to other baselines and even exceeds the performance of FullKV. In specific tasks, EdgeInfinite demonstrates relatively better performance in summarization and code completion, and achieves notable superior results in multi-document QA and few-shot learning.

It can be revealed that KV cache optimization methods generally perform similarly to or slightly better than FullKV. However, EdgeInfinite significantly outperforms FullKV in certain tasks, such as HotpotQA and TriviaQA. The performance enhancement is attributed to its strategy of segmenting long context sequences into multiple shorter sequences, reducing performance degradation from processing excessively long sequences. Meanwhile, EdgeInfinite shows relatively weaker performance in single-document QA than in multi-document QA. This is because single-document QA requires precise answers, while multi-document QA focuses on summarizing content. The memory compression in EdgeInfinite leads to precision loss in KV states, making it better suited for generating summary answers rather than precise retrieval.

\subsection{Ablation Study}

\begin{table}
\centering
\resizebox{\columnwidth}{!}
{
\fontsize{8}{10}\selectfont
\setlength{\tabcolsep}{2pt} 

\begin{tabular}{lccccccc}
\toprule
 & \textbf{SQA} & \textbf{MQA} & \textbf{Sum} & \textbf{FS}  & \textbf{Syn} & \textbf{Code} & \textbf{Avg} \\ 
\midrule
EdgeInfinite & \textbf{23.50} & 24.17 & \textbf{12.10} & \textbf{47.89} & \textbf{11.17} & 37.88 & \textbf{25.71} \\
wo window tokens & 23.28 & \textbf{24.36} & 11.74 & 46.44 & 10.00 & 37.74 & 25.18 \\
wo sink tokens & 20.43 & 19.12 & 11.60 & 43.78 & 6.12 & \textbf{44.28} & 23.17 \\
wo sink \& window & 19.06 & 18.56 & 11.19 & 40.10 & 5.92 & 43.19 & 21.90 \\
\bottomrule
\end{tabular}
}
\caption{\label{Ablation}
Ablation experiment results on LongBench (SQA = Single-Document QA, MQA = Multi-Document QA, Sum = Summarization, FS = Few-shot Learning, Syn = Synthetic).}
\end{table}

To evaluate the impact of retaining specific KV cache during the inference process of EdgeInfinite, we conduct ablation studies to assess the effects of sink tokens and window tokens on inference performance. These ablation experiments are also performed on LongBench. Table~\ref{Ablation} presents the average scores for different task categories and the overall average score under three conditions: removing sink tokens, removing window tokens, and removing both sink and window tokens. 

Removing sink tokens significantly impacts the results of most tasks, as the initial tokens often contain important positional and semantic information for many tasks. Additionally, removing window tokens also affects overall performance. Retaining a fixed number of window tokens avoids the issue of $L_{\mathrm{res}}$ being too short, which would result in too few tokens retained as KV cache at the end of the sequence during memory compression. This mechanism effectively maintains semantic continuity during inference.

\subsection{Efficiency}

\begin{figure}[htbp]
  \includegraphics[width=\columnwidth]{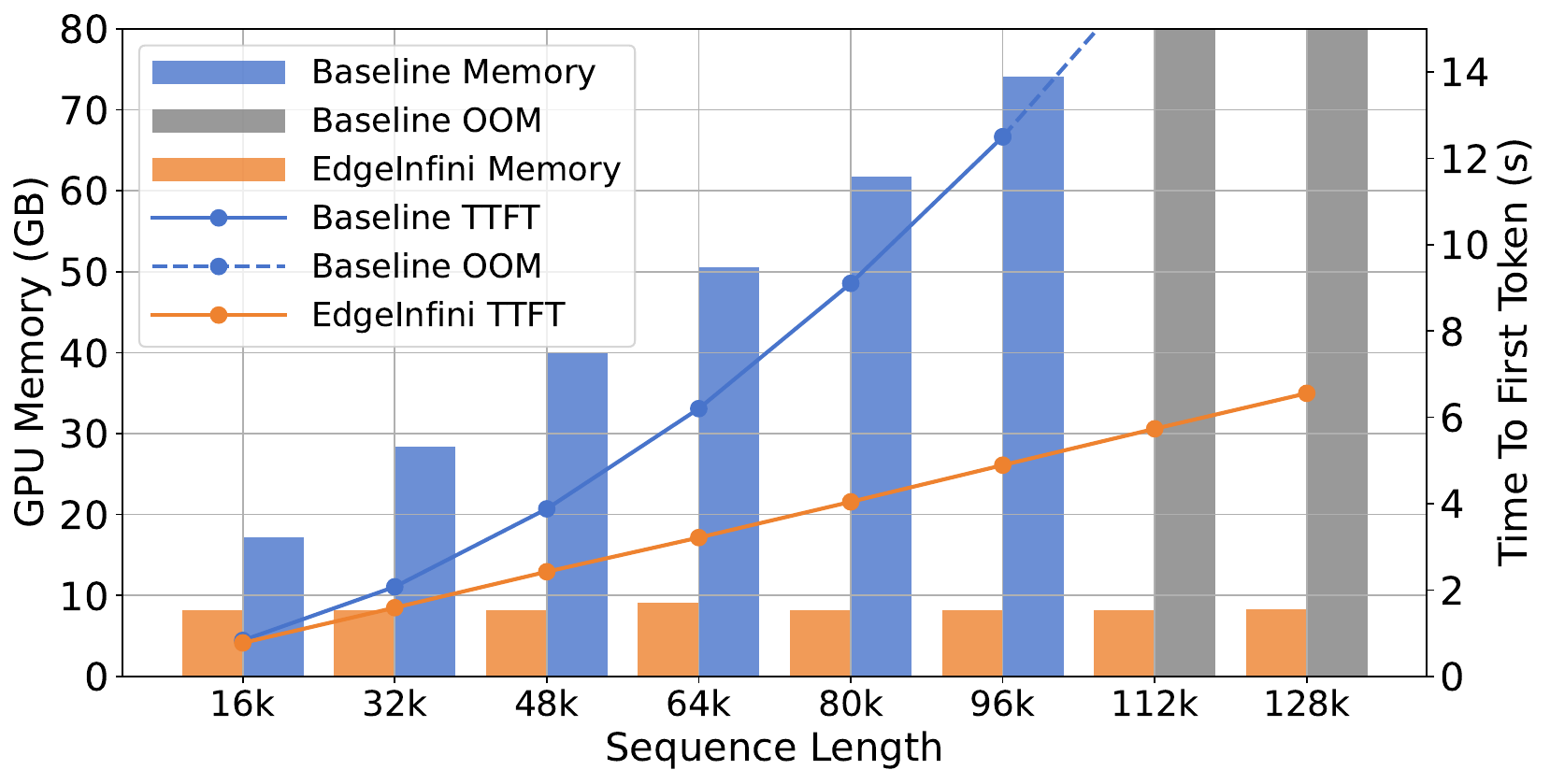}
  \caption{Efficiency of EdgeInfinite. We demonstrate GPU memory consumption and TTFT for varying input sequence lengths.}
  \label{fig:efficiency}
\end{figure}

We compare TTFT and memory usage between EdgeInfinite and the original BlueLM-3B model, as shown in Figure~\ref{fig:efficiency}. The results demonstrate that EdgeInfinite exhibits significant advantages in handling long sequences, with resource consumption not increasing rapidly with sequence length. This is attributed to our method's ability to process long sequences in chunks within the segment size, thereby substantially reducing the computational resource requirements.

\section{Conclusion}

In this study, we propose EdgeInfinite, an efficient method for long context tasks on edge devices. By integrating compressed memory into the Transformer-based LLMs with a trainable memory-gating module, we enable efficient inference on infinite context while maintaining compatibility with the vanilla Transformer architecture. Additionally, we design an effective strategy to retain important tokens during inference for long context tasks to enhance the inference performance, and switch to the original backbone model for short context tasks. Our evaluation on long context benchmarks reveals that EdgeInfinite achieves performance comparable to baseline methods. In summary, EdgeInfinite offers an efficient solution for long context tasks on resource-constrained edge devices.

\bibliography{custom}

\appendix



\end{document}